%% file: main.tex
\documentclass[10pt,twocolumn,letterpaper]{article}

\usepackage[pagenumbers]{iccv} 


\usepackage{graphicx}
\usepackage{booktabs}

\usepackage{url}
\usepackage{algorithm}
\usepackage{algorithmic}
\usepackage{multirow}%
\usepackage{multicol}
\usepackage{amsmath,amssymb,amsfonts}%
\usepackage{amsthm}%
\usepackage{mathrsfs}%
\usepackage[title]{appendix}%
\usepackage{xcolor}%
\usepackage{textcomp}%
\usepackage{manyfoot}%
\usepackage{listings}%
\usepackage{anyfontsize}
\usepackage{color}
\usepackage{bbding}
\usepackage{longtable}
\usepackage{supertabular}
\usepackage{colortbl}
\usepackage{tocloft}
\usepackage{wrapfig} 
\usepackage{mathrsfs}
\usepackage{pifont}

\definecolor{iccvblue}{rgb}{0.21,0.49,0.74}
\usepackage[pagebackref,breaklinks,colorlinks,allcolors=iccvblue]{hyperref}

\def\paperID{7252} 
\def\confName{ICCV}
\def\confYear{2025}

\title{Why Can Accurate Models Be Learned from Inaccurate Annotations?}

\author{Chongjie Si$^1$, Yidan Cui$^1$, Fuchao Yang$^2$, Xiaokang Yang$^1$, Wei Shen$^1$\\
$^1$Shanghai Jiao Tong University, $^2$Southeast University\\
{\tt\small \{chongjiesi, wei.shen\}@sjtu.edu.cn}
}

\begin{document}

\maketitle

\input{sec/0.abstract}

\input{sec/1.intro}

\input{sec/3.exploration}

\input{sec/4.proposed_method}

\input{sec/5.experiments}

\input{sec/2.related_work}

\input{sec/6.conclusion}
{
    \small
    \bibliographystyle{ieeenat_fullname}
    \bibliography{main}
}

\end{document}


\maketitle


\section{Discussion on Sec. 3.2}

In addition to the aforementioned analyses, we can draw several other conclusions or insights.
\begin{itemize}
    
    \item The theoretical analysis allows us to explain a common trend observed in many studies \cite{lv2020progressiveproden,xu2021instancevalen, he2022partialplls, wu2022revisiting, lyu2022deep}: as label inaccuracy increases, there is often a precipitous decline in performance.
    This occurs because when $\rho$ is too large, the angle of the principal subspace may undergo significant changes, resulting in a noticeable shift of the principal subspace. This shift entails a loss of crucial information for the task.
    \item We can posit that the reason why various information and multiple techniques such as regularization or data normalization aid in performance is that they can lower the upper bound of $\sin \theta$, thus preventing significant shifts in the principal subspace.
\end{itemize}

\section{Details on the Baselines}
We provide the details on the baselines here.
\begin{itemize}
    \item PL-CL \cite{si2024partial}: uses the non-candidate labels to induce a complementary classifier to eliminate the false-positive labels [suggested configuration: $k$=10, $\lambda$=0.03, $\gamma$, $\mu$, $\alpha$, and $\beta \in$ \{0.001, 0.01, 0.1, 0.2, 0.5, 1, 1.5, 2, 4\}];

    \item PL-CGR \cite{zhang2024partial}: a cost-guided retraining strategy that guides and corrects disambiguation results while addressing instance-level class imbalance for candidate labels [suggested configuration: $\mu \in$ \{0.04, 0.05\}, $\lambda \in$ \{0.4, 0.5, 0.6\}, $c$=0.2];

    \item DPCLS \cite{jia2023partial}: leverages dissimilarity propagation to guide the shrinkage of candidate labels [suggested configuration: $k$=10, $\lambda$=0.05, $\beta$=0.001, $\alpha \in$ \{0.001, 0.01\}];
    
    \item PL-AGGD \cite{wang2021adaptivePLAGGD}: a approach which constructs a similarity graph and further uses that graph to realize disambiguation [suggested configuration: $k$=10, $T$=20, $\lambda$=1, $\mu$=1, $\gamma$=0.05];
    
    \item SURE \cite{feng2019partialsssfagdsg}: a self-guided retraining based approach which maximizes an infinity norm regularization on the modeling outputs to realize disambiguation [suggested configuration: $\lambda$ and $\beta$ $\in$ \{0.001, 0.01, 0.05, 0.1, 0.3, 0.5, 1\}];
    
    \item LALO \cite{2018Leveraginglalo}: an identification-based approach with constrained local consistency to differentiate the candidate labels [suggested configuration: $k$=10, $\lambda$=0.05, $\mu$=0.005];
    
    \item PL-LEAF \cite{zhang2016partial23232}: a \textit{feature-aware disambiguation} based approach that utilizes the feature manifold to generate labeling confidence and perform multi-output regression to train the predictive model [suggested configuration: $K$=10, $C_1$=10, $C_2$=1];

    \item PICO \cite{wang2022pico}: a deep-learning framework which utilizes a contrastive learning module and a prototype-based label disambiguation algorithm to identify the ground truth;

    \item PRODEN \cite{lv2020progressiveproden}: a deep-learning framework which minimizes a proposed estimator of classification to update model and identify true labels.


    \item PLS \cite{albert2023your}: a deep-learning framework that utilizes a state-of-the-art noise detection metric to identify noisy samples and estimate their true label through a consistency regularization method.

    \item AGCE \cite{zhou2023asymmetric}: a deep-learning framework which presents asymmetric loss functions, enabling the training of a noise-tolerant classifier using noisy labels, provided that clean labels are in the majority.

\end{itemize}

\section{Details on the Real-world Data Sets}

The characteristics of the real-world data sets in Table 1 in the main text are shown in Tab. \ref{tab:real-world data character}.

Specifically, for facial age estimation task, the ages annotated by crowd-sourced labelers are considered as each human face's candidate labels. For automatic face naming task, each face scratched from a video or an image is presented as a sample while the names extracted from the corresponding titles or captions are its candidate labels. For object classification task, image segmentations are taken as instances with objects appearing in the same image as candidate labels.

\begin{table*}[t]\small
\setlength{\tabcolsep}{4.5mm}
\renewcommand\arraystretch{1.0} 
    \caption{Characteristics of the real-world data sets.}

    \centering
    \begin{tabular}{c c c c c c}
     \toprule
        Data set & \#examples & \#features & \#labels & average \#labels & Task Domain\\

        \midrule
      
        FG-NET \cite{panis2016overviewFg-net} & 1002 & 262 & 78 & 7.48 & facial age estimation\\
         Lost \cite{cour2009learningLost} & 1122 & 108 & 16 & 2.23 & automatic face naming\\
        MSRCv2 \cite{liu2012conditional} & 1758 & 48 & 23 & 3.16 & object classification\\
        Mirflickr \cite{huiskes2008mirMirflickr} & 2780 & 1536 & 14 & 2.76 & web image classification \\
        Soccer Player \cite{zeng2013learningSoccerplayer} & 17472 & 279 & 171 & 2.09 & automatic face naming\\
        Yahoo!News \cite{guillaumin2010multipleYahoonews} & 22991 & 163 & 219 & 1.91 & automatic face naming \\
        \bottomrule
    \end{tabular}
    \label{tab:real-world data character}
\end{table*}




      



\section{More Experiment Results}

\subsection{Ablation Study on PRODEN}
We present the ablation study on PRODEN in Table. \ref{tab:ablation study2}.

\begin{table*}[!ht]
\renewcommand\arraystretch{0.9} 
    \centering
    \caption{Ablation study of LIP coupled with PRODEN. }
    \resizebox{\textwidth}{!}{
    \begin{tabular}{c c l l l l l l}\toprule
        \multirow{2}{*}{PSP} & \multirow{2}{*}{LAP} & \multicolumn{3}{c}{CIFAR-100} & \multicolumn{3}{c}{CUB-200} \\ \cline{3-8} & & \multicolumn{1}{c}{ $p=0.05$} & \multicolumn{1}{c}{$p=0.1$} & \multicolumn{1}{c}{$p=0.2$} & \multicolumn{1}{c}{$p=0.02$} & \multicolumn{1}{c}{$p=0.04$} &
         \multicolumn{1}{c}{$p=0.06$} \\ 
         
         \midrule

          \ding{53} & \ding{53} & 77.12 $\pm$ 0.13\% & 76.04 $\pm$ 0.16\% & 56.83 $\pm$ 0.05\% & 74.53 $\pm$ 0.03\% & 74.36 $\pm$ 0.15\% & 72.01 $\pm$ 0.18\% \\ 

          \checkmark & \ding{53} & 77.21 $\pm$ 0.01\% & 76.25 $\pm$ 0.15\% & 56.89 $\pm$ 0.01\% & 74.66 $\pm$ 0.01\% & 74.41 $\pm$ 0.07\% & 72.17 $\pm$ 0.04\%  \\

         \checkmark & \checkmark & 77.25 $\pm$ 0.03\% & 76.42 $\pm$ 0.05\% & 56.90 $\pm$ 0.01\% & 74.73 $\pm$ 0.01\% & 74.51 $\pm$ 0.02\% & 72.33 $\pm$ 0.03\% \\
       
         \bottomrule
    \end{tabular}
    }       
    \label{tab:ablation study2}
\end{table*}

{
    \small
    \bibliographystyle{ieeenat_fullname}
    \bibliography{main}
}

%% file: sec/0.abstract.tex
\begin{abstract}

Learning from inaccurate annotations has gained significant attention due to the high cost of precise labeling. 
However, despite the presence of erroneous labels, models trained on noisy data often retain the ability to make accurate predictions.
This intriguing phenomenon raises a fundamental yet largely unexplored question: why models can still extract correct label information from inaccurate annotations remains unexplored. 
In this paper, we conduct a comprehensive investigation into this issue. 
By analyzing weight matrices from both empirical and theoretical perspectives, we find that label inaccuracy primarily accumulates noise in lower singular components and subtly perturbs the principal subspace. 
Within a certain range, the principal subspaces of weights trained on inaccurate labels remain largely aligned with those learned from clean labels, preserving essential task-relevant information. 
We formally prove that the angles of principal subspaces exhibit minimal deviation under moderate label inaccuracy, explaining why models can still generalize effectively.
Building on these insights, we propose LIP, a lightweight plug-in designed to help classifiers retain principal subspace information while mitigating noise induced by label inaccuracy. 
Extensive experiments on tasks with various inaccuracy conditions demonstrate that LIP consistently enhances the performance of existing algorithms.
We hope our findings can offer valuable theoretical and practical insights to understand of model robustness under inaccurate supervision.

\end{abstract}

%% file: sec/1.intro.tex
\section{Introduction}
\label{sec:intro}

Inaccurate annotations are prevalent in real-world scenarios, arising from the prohibitive costs and challenges associated with obtaining perfectly labeled data \cite{jia2023complementary, zhou2018brief, zhu2009introductionddd, si2023appeal}.
Despite the inherent inaccuracies in annotations, an intriguing phenomenon emerges: models trained on such noisy data often retain the ability to make correct predictions, sometimes even surpassing those trained with ground-truth labels in certain cases. This paradox raises a fundamental question that remains unresolved: \textbf{Why can models extract correct label information from inaccurate annotations?}

In this paper, we conduct a comprehensive investigation into this issue.
Generally, models encapsulate the knowledge learned during training within their weight matrices. 
We first investigate how varying degrees of label inaccuracy affect the weight matrix, both from the perspective of singular values and subspace structure. 
Our findings show that large label inaccuracy accumulates noise in the lower singular components and also alters the structure of the principal subspace, leading to the loss of critical task-specific information.
However, \textbf{within a certain range, label inaccuracy does not significantly affect, or may even leave unchanged, the principal subspace of the weight matrix}.
Subsequently, we provide a theoretical proof to explain the aforementioned phenomenon. 
We find that within a certain range of label inaccuracy, the angles of the principal subspaces do not exhibit great variation. 
This ultimately leads us to conclude that models can learn correct information from inaccurate labels because \textbf{the principal subspaces of the weights trained on inaccurate labels do not significantly deviate from those derived from accurate labels.}

Furthermore, based on our findings, we propose a plug-in named LIP, as ``Label Inaccuracy Processor''. 
It aids existing classifiers in retaining information of principal subspaces while reducing the noises caused by label inaccuracy. 
This plug-in is lightweight and has been demonstrated through extensive experiments under various configurations of label inaccuracy to effectively enhance the performance of existing algorithms.
The main contributions of this paper are as follows:

\begin{itemize}
    \item To the best of our knowledge, it is the first investigation into why model can extract correct label information from inaccurate annotations. We provide both empirical and theoretical proof on this question, which may guide the future exploration of the related areas.
    \item We propose a lightweight but effective plug-in LIP, which helps existing classifiers retain core information and filter the noises caused by label inaccuracy.
    \item We conducted extensive evaluations across multiple algorithms on tasks with different inaccurate label settings, demonstrating the clear superiority of our approach.
\end{itemize}

%% file: sec/3.exploration.tex
\section{Extracting Accurate Labels from Inaccurate Annotations: Empirical Evidence}\label{sec empirical}

As stated in the introduction, it remains unclear why models can learn correct labels from inaccurate annotations. 
In other words, we lack a fundamental understanding of what enables models to possess this capability.
In this section, we will systematically explore this issue, beginning with empirical evidence.
We first establish some foundational concepts. 
Generally, models store the critical information for classification tasks captured from training data in their final classification weights. 
For stand-alone methods, these weights are typically obtained by solving an optimization problem. 
For deep-learning-based methods, these weights correspond to the final fully connected (FC) layer used for classification. 
We denote the weights learned from clean labels as $\mathbf{W}$, and those learned from inaccurate labels as $\mathbf{W}'$.

Indeed, exploring why it is achievable is quite challenging and somewhat clueless, akin to questioning why apples fall from trees, a mystery that perplexed thinkers before Newton’s elucidation of gravity \cite{guicciardini2005isaac}.
We know that they do, but we may not fully understand why it happens this way.
However, we are not entirely without clues: to investigate why it is achievable, it is crucial to understand \textbf{how label inaccuracy impacts the weight matrix}. 
Therefore, we believe that an exploration of the relationship between $\mathbf{W}$ and $\mathbf{W}'$ may provide valuable insights into this process.
Given that $\mathbf{W}'$, learned from inaccurate labels, also supports classification tasks and often achieves accuracy close to that of $\mathbf{W}$ \cite{yi2019probabilistic,si2024partial,wang2022pico,lv2020progressiveproden}, it suggests an inherent connection between them. 
What, then, is the nature of this relationship?

\subsection{Observations on the Singular Values}

We first train a well-established classifier, PRODEN \cite{lv2020progressiveproden}, which utilizes a ResNet-34 \cite{he2016deep} backbone, on the clean CIFAR-100 dataset \cite{krizhevsky2009learning}.
This process yields the classification weights $\mathbf{W}$ of the FC layer. 
To introduce varying levels of label inaccuracy, we randomly flip the labels of each sample with a predefined probability $p$, generating datasets with different degrees of annotation noise. 
This procedure results in two possible scenarios: a sample may acquire additional incorrect labels alongside its true label, or it may entirely lose its correct label.
We also enforce a constraint that each sample retains at least one label.
We set $p$ to 0.05, 0.1, 0.2, 0.3, and 0.4, and train the same model on these noisy datasets, yielding the corresponding classification weights $\mathbf{W}'$ for each level of label inaccuracy.

\begin{figure}
    \centering
    \includegraphics[width=1\linewidth]{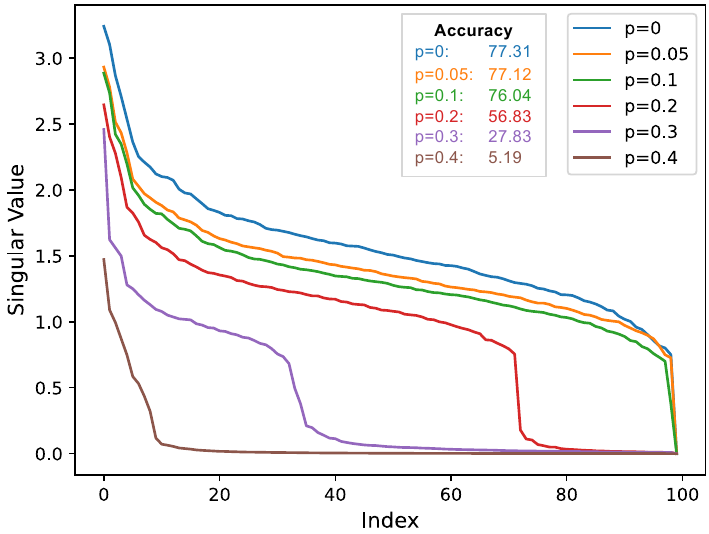}
    \caption{Singular values under different label inaccuracy (i.e., $p$). We also provide the corresponding classification accuracy.}
    \label{fig: singular value}
\end{figure}

To begin with, we perform Singular Value Decomposition (SVD) on both matrices $\mathbf{W}$ and $\mathbf{W}'$ to extract their corresponding singular values $\mathbf{\Sigma}$ and $\mathbf{\Sigma}$' and right singular unitary matrices, referred to as $\mathbf{V}$ and $\mathbf{V}'$. 
We then analyze the variation in singular values of different weights as inaccuracy increased, and the results are shown in the Fig. \ref{fig: singular value}.
First, we observe that the top singular values of the weight matrix do not show significant differences within a certain range of label inaccuracy.
However, once the label inaccuracy exceeds a certain threshold, the top singular values decline rapidly. 
This indicates a significant reduction in information accumulation along the principal directions of the weight matrix, suggesting that the model’s ability to capture key features is diminished.
As noise and uncertainty in the labels increase, these disturbances disrupt the primary structure of the weight matrix, causing the singular value spectrum to flatten.
This flattening also reflects a decrease in the model’s effective rank, weakening the information retained along the principal directions.

Moreover, increasing label inaccuracy accelerates the decay of singular values, as noise predominantly corrupts lower-rank directions, inhibiting meaningful feature accumulation.
When inaccuracy becomes severe, noise overwhelms these subspaces, causing the smallest singular values to rapidly approach zero.
This further diminishes the weight matrix’s effective rank, rendering the model increasingly unstable and less reliable in these directions.

\begin{figure*}
    \centering
    \includegraphics[width=0.8\linewidth]{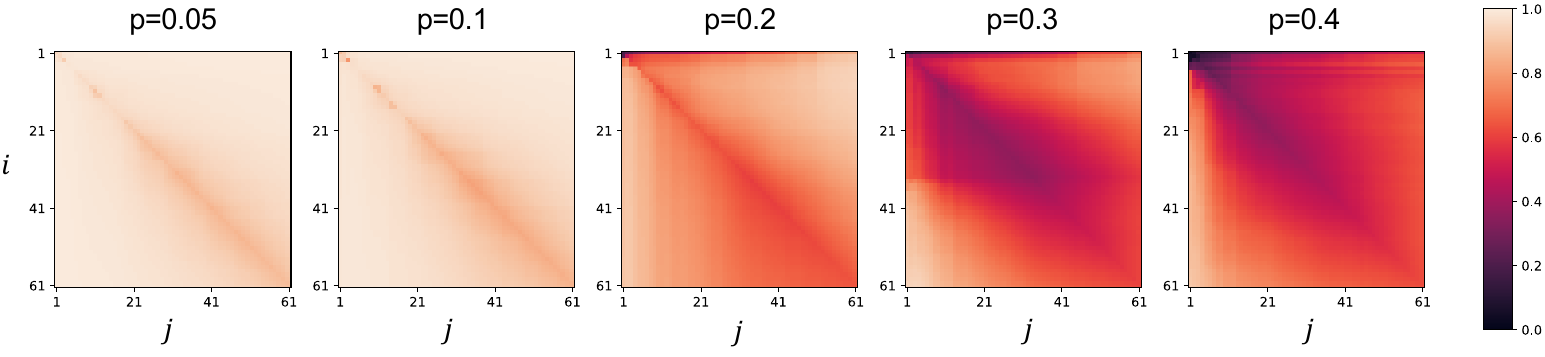}
    \caption{Subspace similarity between top-$i$ column vectors of $\mathbf{W}$ and top-$j$ of $\mathbf{W}'$. As label inaccuracy increases, the weight matrix is able to preserve the characteristics of the principal subspace within a certain range of inaccuracy, thus retaining the most important task-specific information. However, beyond this range, the principal subspace of the weights begins to change, and even eventually becomes completely uncorrelated, leading to the loss of critical information. For clearer illustrations, we present the analysis for $i,j \in [1,60]$.}
    \label{fig:subspace}
\end{figure*}

While the analysis of singular values provided important insights into how label inaccuracy affects the weight matrix, it leaves some critical aspects unclear.
Specifically, while we can see the reduction in the rank and the weakened information retention in both primary and other directions, it remains ambiguous why the model is still capable of learning from inaccurate labels. 
In other words, we do not yet fully understand how much of the essential task-specific information is preserved in the weight matrix despite the presence of label inaccuracy. 
To further explore this, we now turn to the subspace similarities between $\mathbf{W}$ and $\mathbf{W}'$.
By examining the subspace structure, we seek to understand how label inaccuracy influences the information encoded in the weight matrix and to identify whether a shared underlying structure enables the extraction of correct information from inaccurate labels.

\subsection{Similarities of Subspace}

We assess the similarities of the subspace spanned by the top-$i$ singular vectors in $\mathbf{V}$ and that of the top-$j$ singular vectors of $\mathbf{V}'$.
Following \cite{hu2021lora}, we compute the normalized subspace similarity based on the Grassmann distance as
\begin{equation}
    \phi(\mathbf{V}, \mathbf{V}', i, j) = \frac{\|\mathbf{V}_{:i}^{\mathsf{T}}\mathbf{V}'_{:j} \|_F^2}{\min (i,j)} \in [0,1].
\end{equation}
Here, $\phi(\cdot)$ ranges from 0 to 1, where 1 indicates a complete overlap of subspaces and 0 signifies total separation. $\mathbf{V}_{:i}$ and $\mathbf{V}'_{:j}$ represent the top-$i$ and top-$j$ column vectors of $\mathbf{V}$ and $\mathbf{V}'$, respectively. 
As shown in Fig. \ref{fig:subspace}, we observed an important phenomenon: as label inaccuracy increases, the overall divergence between the subspaces of $\mathbf{V}$ and $\mathbf{V}'$ becomes more pronounced, with the most notable changes in the \textbf{principal subspace} (corresponding to the largest singular values).
Since the top singular components capture the most crucial task-specific information, changes in the principal subspace signify the loss of important information, leading to weaker de-noise ability, which can be validated by the classification accuracy in Fig. \ref{fig: singular value}.
Therefore, label inaccuracy not only introduces noise in the lower singular components, but also disrupts the principal subspace, leading to the loss of key information.

However, surprisingly, within a certain range of label inaccuracy ($p \leq 0.1$ in our experiments), the principal subspace of the weights remains largely unaffected and even nearly identical. 
As seen in Fig. \ref{fig: singular value}, the classification results of $\mathbf{W}'$ under this level of inaccuracy are nearly identical to those of $\mathbf{W}$. 
This similarity between the principal subspaces of $\mathbf{V}$ and $\mathbf{V}'$ indicates that $\mathbf{W}'$ effectively preserves task-specific information, as if it were learned from clean data.
\textbf{This may be a key reason why the model can still effectively learn from inaccurate annotations.}

It appears we are moving closer to the truth. However, to truly grasp the essence, we must address a pivotal question: Why are their principal subspaces so similar in a certain range of label inaccuracy?

\section{Theoretical Analysis} \label{sec theorectical}

\subsection{Label Inaccuracy can be Viewed as a Form of Weight Perturbation}

Formally speaking, suppose $\mathcal{X}=\mathbb{R}^{q}$ denotes the $q$-dimensional feature space, and $\mathcal{Y}=\{0,1\}^{l}$ is the label space with $l$ classes.
Given an inaccurate label data set $\mathcal{D} = \{\mathbf{x}_{i},\mathcal{S}_{i} | 1\leq i \leq n \}$, where $\mathbf{x}_{i} \in \mathcal{X}$ is the $i$-th sample and $\mathcal{S}_{i} \subseteq \mathcal{Y}$ is the corresponding corrupted label set, we learn a classifier $f:\mathcal{X}\rightarrow \mathcal{Y}$ based on $\mathcal{D}$. 
Suppose $\mathbf{X} = [\mathbf{x}_1,\mathbf{x}_2,\dots,\mathbf{x}_{n}]^{\mathsf{T}} \in \mathbb{R}^{n \times q}$ denote the $q$-dimensional instance matrix with $n$ instances, and $\mathbf{G} = [\mathbf{g}_1,\mathbf{g}_2,\dots,\mathbf{g}_{n}]^{\mathsf{T}} \in \{0,1\}^{n\times l}$ represents the ground-truth label matrix. We aim to learn a weight matrix $\mathbf{W}\in\mathbb{R}^{q\times l}$ to map the instance matrix to the ground-truth one. Consider the following simple classifier (or loss function):
\begin{equation}
    \min_\mathbf{W} \| \mathbf{X}\mathbf{W} - \mathbf{G} \|_F^2 + \lambda\|\mathbf{W}\|_F^2,
    \label{eq simple optimization}
\end{equation}
which is a regularized least squares problem, with $\lambda$ as a hyper-parameter. The closed-form solution is 
\begin{equation}
    \mathbf{W} = (\mathbf{X}^\mathsf{T}\mathbf{X} + \lambda\mathbf{I})^{-1} \mathbf{X}^\mathsf{T} \mathbf{G},
\end{equation}
where $\mathbf{I}$ is the identity matrix. For convenience, we denote $\mathbf{K} = \mathbf{X}^\mathsf{T}\mathbf{X} + \lambda \mathbf{I}$ to simplify following formula expressions.

When handling inaccurate labels, we define the corrupted label set $\mathbf{Y} = [\mathbf{y}_1,\mathbf{y}_2,\dots,\mathbf{y}_{n}]^{\mathsf{T}} \in \{0,1\}^{n\times l}$, where $y_{ij}=1$ if the $j$-th label of $\mathbf{x}_{i}$ is annotated as its ``ground-truth'' label.
Indeed, we can consider $\mathbf{Y}$ as a perturbation of $\mathbf{G}$, where $\mathbf{Y} = \mathbf{G} + \mathbf{M}$. 
Here, $\mathbf{M}$ is a perturbation mask composed of elements from ${-1,0,1}$, representing the label inaccuracy. 
To ensure label consistency, we impose the constraint $\mathbf{M} + \mathbf{G} \geq 0$, meaning that ${M}_{ij}$ can take the value -1 only where ${G}_{ij} = 1$, ensuring that incorrectly removed labels originate from initially assigned ones.
Therefore, the learned weight matrix $\mathbf{W}'$ under supervision of $\mathbf{Y}$ is 
\begin{align}
     \mathbf{W}' & = \mathbf{K}^{-1} \mathbf{X}^\mathsf{T} \mathbf{Y} =  \mathbf{K}^{-1} \mathbf{X}^\mathsf{T} \mathbf{G} + \mathbf{K}^{-1} \mathbf{X}^\mathsf{T} \mathbf{M} \\ \nonumber
     & = \mathbf{W} + \Delta \mathbf{W}.
\end{align}
Consequently, the impact of label inaccuracy on the weight $\mathbf{W}$ can be described as introducing an additional perturbation, denoted as $\Delta \mathbf{W} = \mathbf{K}^{-1} \mathbf{X}^\mathsf{T} \mathbf{M}$.

\subsection{Estimation of Upper Bound of Perturbation}

We then calculate the upper bound of $\Delta \mathbf{W}$. According to norm inequalities \cite{petersen2008matrix}, we have
\begin{equation}
\begin{aligned}
     \|\Delta \mathbf{W}\|_F & \leq \|\mathbf{K}^{-1} \mathbf{X}^\mathsf{T}\|_2  \|\mathbf{M}\|_F \\
     & \leq \|\mathbf{K}^{-1}\|_2 \|\mathbf{X}^\mathsf{T}\|_2 \|\mathbf{M}\|_F,
\end{aligned}
\end{equation}
where $\|\cdot\|_F$ represents the Frobenius norm and $\|\cdot\|_2$ denotes the spectral norm of a matrix, respectively. Denote $\lambda_{\min}(\cdot)$ as the minimum eigenvalue of a matrix, and $\sigma_{\max} (\cdot)$ as the maximum singular value of a matrix. Due to $q \ll n$ generally, $\mathbf{X}$ is expected to be full rank. We therefore have
\begin{equation}
\begin{aligned}
    \| \mathbf{K} ^{-1} \|_2 & = \| (\mathbf{X}^\mathsf{T}\mathbf{X} + \lambda \mathbf{I})^{-1} \|_2 = \frac{1}{\lambda_{\min}(\mathbf{X}^\mathsf{T}\mathbf{X}) + \lambda} \\
    \| \mathbf{X}^\mathsf{T} \|_2  & = \| \mathbf{X} \|_2 = \sigma_{\max}(\mathbf{X}).
\end{aligned}
\end{equation}
For simplicity, we define the degree of label inaccuracy as $p$, where $P(M_{ij} \neq 0) = p$. Consequently, the total number of non-zero elements in $\mathbf{M}$ is $p n l$, and we have $\|\mathbf{M}\|_F = \sqrt{p n l} $. Therefore, the upper bound of $\Delta\mathbf{W}$ is given by
\begin{equation}
    \|\Delta\mathbf{W}\|_F \leq \frac{\sigma_{\max}(\mathbf{X}) \sqrt{nl}}{\lambda_{\min}(\mathbf{X}^\mathsf{T}\mathbf{X}) + \lambda} \sqrt{p}.
\end{equation}

\begin{figure*}[!ht]
    \centering
    \includegraphics[width=0.9\linewidth]{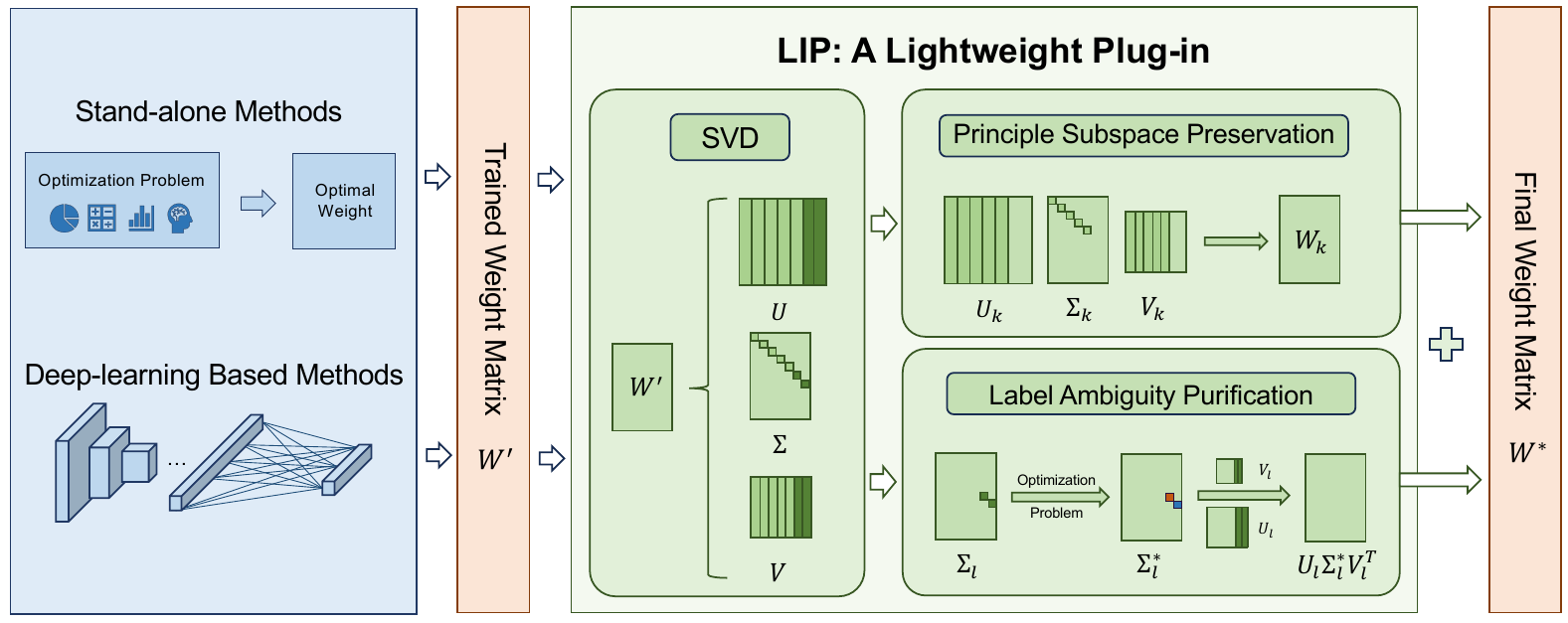}
    \caption{Framework of LIP. Once a method is trained, the LIP applies post-processing to the trained weights used for classification. The two modules of the LIP, Principle Subspace Preservation and Label Ambiguity Purification, are utilized to retain critical information in the principal subspace and to rectify noise resulting from label inaccuracy, respectively. Given the high processing speed and extreme lightness of these modules, coupled with the absence of a need for training, LIP emerges as an efficient and lightweight plug-in.}
    \label{fig: framework}
\end{figure*}

\subsection{Subspace Perturbation Analysis}

To assess the impact of perturbation on the subspace, we apply the Davis-Kahan sine theorem \cite{davis1970rotation}, which provides an upper bound on the change in the principal angles between the perturbed and unperturbed subspaces. Specifically, the theorem states that for a matrix perturbation $\Delta \mathbf{W}$, the sine of the angle $\theta$ between the subspaces of $\mathbf{W}$  and $\mathbf{W}'$ is bounded by
\begin{equation}
    \sin\theta \leq \frac{\| \Delta \mathbf{W} \|_2}{\delta},
\end{equation}
where $\delta$ denotes the gap between the principle singular values of $\mathbf{W}$ (it can be viewed as a hyper-parameter). Since $\|\Delta\mathbf{W}\|_2 \leq \| \Delta\mathbf{W}\|_F$ \cite{petersen2008matrix}, the bound becomes
\begin{equation}
    \sin\theta \leq \frac{\sigma_{\max}(\mathbf{X}) \sqrt{nl}}{\delta(\lambda_{\min}(\mathbf{X}^\mathsf{T}\mathbf{X}) + \lambda)} \sqrt{p}.
\end{equation}
From this inequality we can draw an important conclusion: when the degree of label inaccuracy $p$ remains within a reasonable range, the angle remains small, and the principal subspace of the weight matrix $\mathbf{W}$ does not undergo significant changes or the changes are controllable. 
This implies that even in the presence of label inaccuracy, classifiers can still learn information crucial for classification that can be learned from clean labels. 
Therefore, through experimental results and theoretical analysis, we ultimately determine why models can learn correct information from inaccurate annotations: 
\begin{center}
    \textit{\textbf{A certain degree of label inaccuracy does not cause the principal subspace of the information learned from clean data to shift severely.}}
\end{center}

%% file: sec/4.proposed_method.tex
\section{LIP: A Lightweight Plug-in}

According to our previous analysis, label inaccuracy within a certain range does not significantly impact the critical information in the principal subspace, but instead primarily introduces noise into the bottom singular components of the weight matrix.
This naturally leads us to contemplate whether a method can be devised that not only preserves the information in the principal subspace but also reduces or even refine the impact of noise introduced by label inaccuracy. 
Based on this consideration, we propose a post-processing plug-in, named LIP. 
LIP primarily consists of two phases: Principle Subspace Preservation (PSP) and Label Ambiguity Purification (LAP). 
The framework of LIP is shown in Fig. \ref{fig: framework}.

\subsection{Principle Subspace Preservation}

Consider the weight matrix $\mathbf{W}'$ learned from datasets with inaccurate labels. 
For a stand-alone method, $\mathbf{W}'$ represents the ultimate optimization goal; for deep-learning-based methods, $\mathbf{W}'$ corresponds to the weights of the final FC layer used for classification. 
As demonstrated in Secs. \ref{sec empirical}-\ref{sec theorectical}, the crucial information for classification using inaccurate labels is preserved within the principal subspace of the weights, i.e., the top singular components of $\mathbf{W}'$. 
Therefore, we initially focus on preserving this information.

We perform SVD on $\mathbf{W}'$ to obtain the corresponding left and right singular vectors $\mathbf{U}$ and $\mathbf{V}$, as well as the singular value matrix $\mathbf{\Sigma} = {\rm diag}(\sigma_1, \sigma_2,\dots, \sigma_{\min(q,l)})$ with $\sigma_i$ arranged in descending order. 
Let $k$ ($k \leq \min(q,l)$) denote the number of top singular components to be retained, which is a hyper-parameter. Let $\mathbf{U}_k = [\mathbf{u}_1,\mathbf{u}_2,\dots,\mathbf{u}_{k}]$, $\mathbf{V}_k = [\mathbf{v}_1,\mathbf{v}_2,\dots,\mathbf{v}_{k}]$ represent the top $k$ singular vectors and $\mathbf{\Sigma}_k = {\rm diag}(\sigma_1, \sigma_2,\dots, \sigma_k)$. Thus, we can express the matrix $\mathbf{W}_k$ which contains the crucial information derived from principle subspace as:

\begin{equation}
     \mathbf{W}_k = \mathbf{U}_k\mathbf{\Sigma}_k\mathbf{V}_k^{\mathsf{T}}.
\end{equation}

\subsection{Label Ambiguity Purification}

The matrix $\mathbf{W}_k$, which encapsulates the principal subspace’s crucial information, has already mitigated the noise introduced by label inaccuracy. 
However, several studies \cite{si2024unleashing,si2024flora,si2024see,han2023svdiff} highlight the importance of the bottom singular components to the weight matrix itself. 
Notably, as illustrated in Fig. \ref{fig: singular value}, these bottom singular values often appear flattened and near-zero due to label inaccuracy, suggesting that label noise may significantly suppress these important components.
Building on this observation, we propose that these components are often misperceived as noise.
Instead of discarding them, we aim to refine these singular values, allowing them to recapture essential information from the training data and reassess the importance of each singular component, thereby enhancing the utilization of the matrix’s effective rank.

Specifically, suppose $\mathbf{U}_{l}=[\mathbf{u}_{k+1},\mathbf{u}_{k+2},\dots,\mathbf{u}_{\min(q,l)}]$ and $\mathbf{V}_{l}=[\mathbf{v}_{k+1},\mathbf{v}_{k+2},\dots,\mathbf{v}_{\min(q,l)}]$ denote the singular vectors not retained in the PSP phase, and $\mathbf{\Sigma}_{l}={\rm diag}(\sigma_{k+1},\sigma_{k+2},\dots,\sigma_{\min(q,l)})$ represents the singular values not preserved. With these definitions, we can express the weight matrix $\mathbf{W}'$ as follows:
\begin{equation}
    \mathbf{W}' = \mathbf{W}_k + \mathbf{U}_{l}\mathbf{\Sigma}_{l}\mathbf{V}_{l}^\mathsf{T}.
\end{equation}
To refine the singular values, we retrain them using the training data to extract critical information, which leads to the following optimization problem:
\begin{equation}
\begin{aligned}
    \min_{\mathbf{\Sigma}_{l}} & \quad \|\mathbf{X}\mathbf{W}' -\mathbf{Y}\|_{F}^2 \\
    {\rm s.t.} & \quad \mathbf{W}' = \mathbf{W}_k + \mathbf{U}_{l}\mathbf{\Sigma}_{l}\mathbf{V}_{l}^\mathsf{T}.
\end{aligned}
\label{eq optimization lap}
\end{equation}
By solving the aforementioned optimization problem, we can determine the optimal singular matrix $\mathbf{\Sigma}_l^*$. Consequently, this enables us to derive the final weight matrix $\mathbf{W}^*$ for classification as
\begin{equation}
    \mathbf{W}^* = \mathbf{W}_k + \mathbf{U}_{l} \mathbf{\Sigma}_{l}^* \mathbf{V}_{l}^\mathsf{T}.
\end{equation}
This matrix not only preserves information in the principal subspace but also effectively addresses and refines the noise associated with label inaccuracy.

\subsection{Numerical Solution to Eq. (\ref{eq optimization lap})}

We first reformulate Eq. (\ref{eq optimization lap}) as
\begin{equation}
    \min_{\mathbf{\Sigma}_{l}}  \quad  \|\mathbf{X}(\mathbf{W}_k + \sum_{i=k+1}^{\min(q,l)} \sigma_i \mathbf{u}_i\mathbf{v}_i^\mathsf{T}) -\mathbf{Y}\|_{F}^2.
    \label{eq optimization full}
\end{equation}
Taking the gradient of Eq. (\ref{eq optimization full}) w.r.t. each $\sigma_j$, we have
\begin{equation}
    \nabla_{\sigma_j} = 2(\sigma_j \mathbf{u}_j^\mathsf{T}\mathbf{X}^\mathsf{T} \mathbf{X}\mathbf{u}_j + \mathbf{u}_j^\mathsf{T}\mathbf{X}^\mathsf{T} (\mathbf{X}\mathbf{W}_k - \mathbf{Y})\mathbf{v}_j ).
\end{equation}
The optimal solution of Eq. (\ref{eq optimization full}) is achieved when $\nabla_{\sigma_j} = 0$, and accordingly we have
\begin{equation}
    \sigma_j = \frac{\mathbf{u}_j^\mathsf{T}\mathbf{X}^\mathsf{T} (\mathbf{Y} - \mathbf{X}\mathbf{W}_k)\mathbf{v}_j}{\mathbf{u}_j^\mathsf{T}\mathbf{X}^\mathsf{T} \mathbf{X}\mathbf{u}_j}.
\end{equation}
Thus, the corresponding closed-form solution to the optimization problem in Eq. (\ref{eq optimization lap}) is:
\begin{equation}
    \mathbf{\Sigma}_{l}^{*} = \frac{ {\rm diag} (\mathbf{U}_l^\mathsf{T} \mathbf{X}^\mathsf{T} (\mathbf{Y} - \mathbf{X}\mathbf{W}_k) \mathbf{V}_l )}{ {\rm diag}(\mathbf{U}_l^\mathsf{T}  \mathbf{X}^\mathsf{T} \mathbf{X} \mathbf{U}_l )}.
    \label{eq closed form solution}
\end{equation}
Given that this problem has a closed-form solution, LIP does not require training (i.e., it is training-free). 
Instead, LIP can be directly integrated into any pre-trained classifier, adding only two computational steps to preserve crucial information and refine noise from label inaccuracy.

\begin{table*}[!ht]
\renewcommand\arraystretch{0.9} 
    \centering
    \caption{Classification accuracy of each compared approach on the real-world data sets. * denotes these datasets are processed with noise. For any baselines, $\bullet$/$\circ$ indicates whether the performance of baseline coupled with LIP is statistically superior/inferior to that of the baseline itself based on pairwise $t$-test at significance level of 0.05. ``$\uparrow$Ratio'' represents the average improvement ratio of each baseline by LIP. }
    \resizebox{\textwidth}{!}{
    \begin{tabular}{c l l l l l l c c  }\toprule
         \multirow{2}{*}{Approaches} & \multicolumn{6}{c}{Data set} & \multirow{2}{*}{Avg.} & \multirow{2}{*}{$\uparrow$Ratio} \\  &\multicolumn{1}{c}{FG-NET*} & \multicolumn{1}{c}{Lost*} & \multicolumn{1}{c}{MSRCv2*} & \multicolumn{1}{c}{Mirflickr*} & \multicolumn{1}{c}{Soccer Player*} &
         \multicolumn{1}{c}{Yahoo!News*} \\
         \midrule

        PL-CL & 6.6 $\pm$ 1.2 & 68.3 $\pm$ 2.4 & 45.1 $\pm$ 1.5 & 63.2 $\pm$ 1.3 & 52.1 $\pm$ 0.5 & 60.9 $\pm$ 0.4 & 49.4 & \multirow{2}{*}{5.7\%} \\
        
        + LIP & 6.9 $\pm$ 1.0 $\bullet$ & 72.8 $\pm$ 1.6 $\bullet$ & 52.1 $\pm$ 2.0 $\bullet$ & 64.1 $\pm$ 1.6 $\bullet$ & 53.6 $\pm$ 0.5 $\bullet$ & 63.9 $\pm$ 0.3 $\bullet$ & \textbf{52.2} \\ 
        
        \midrule

        PL-CGR & 5.9 $\pm$ 1.3 & 71.3 $\pm$ 3.0 & 49.7 $\pm$ 2.3 & 64.5 $\pm$ 1.6 & 53.3 $\pm$ 0.9 & 61.6 $\pm$ 0.5 & 51.1 & \multirow{2}{*}{4.6\%} \\
        
        +LIP & 7.9 $\pm$ 1.3 $\bullet$ & 74.3 $\pm$ 2.8 $\bullet$ & 51.6 $\pm$ 2.3 $\bullet$ & 65.8 $\pm$ 1.8 $\bullet$ & 55.9 $\pm$ 0.8 $\bullet$ & 65.1 $\pm$ 0.3 $\bullet$ & \textbf{53.4} \\ \midrule

        DPCLS & 5.7 $\pm$ 0.7 & 69.6 $\pm$ 1.8 & 53.6 $\pm$ 1.4 & 61.2 $\pm$ 1.6 & 54.7 $\pm$ 0.7 & 61.9 $\pm$ 0.4 & 51.1 & \multirow{2}{*}{3.3\%} \\
        
        +LIP & 7.8 $\pm$ 1.0 $\bullet$ & 72.7 $\pm$ 1.8 $\bullet$ & 54.5 $\pm$ 1.4 $\bullet$ & 62.3 $\pm$ 1.6 $\bullet$ & 55.7 $\pm$ 0.6 $\bullet$ & 63.8 $\pm$ 0.3 $\bullet$ & \textbf{52.8} \\ \midrule
    
        PL-AGGD & 6.3 $\pm$ 0.9 & 68.2 $\pm$ 3.0 & 45.0 $\pm$ 1.5 & 63.7 $\pm$ 1.2 & 51.7 $\pm$ 0.5 & 60.7 $\pm$ 0.2 & 49.3 & \multirow{2}{*}{6.3\%} \\ 
        
        + LIP & 6.7 $\pm$ 0.7 $\bullet$ & 72.9 $\pm$ 2.5 $\bullet$ & 52.3 $\pm$ 1.8 $\bullet$ & 64.8 $\pm$ 1.4 $\bullet$ & 53.6 $\pm$ 0.6 $\bullet$ & 63.9 $\pm$ 0.3 $\bullet$ & \textbf{52.4} \\ \midrule
     
        SURE & 5.4 $\pm$ 1.1 & 68.3 $\pm$ 2.0 & 43.8 $\pm$ 2.3 & 61.4 $\pm$ 1.1 & 51.2 $\pm$ 0.5 & 59.1 $\pm$ 0.3 & 48.2 & \multirow{2}{*}{3.7\%}\\
        
        + LIP & 5.9 $\pm$ 1.0 $\bullet$ & 70.6 $\pm$ 1.8 $\bullet$ & 48.1 $\pm$ 2.1 $\bullet$ & 62.2 $\pm$ 1.2 $\bullet$ & 51.8 $\pm$ 0.4 $\bullet$ & 61.6 $\pm$ 0.3 $\bullet$ & \textbf{50.0} \\ \midrule
    
        LALO & 6.0 $\pm$ 0.8 & 66.7 $\pm$ 2.0 & 44.1 $\pm$ 1.7 & 62.1 $\pm$ 1.5 & 51.5 $\pm$ 0.4 & 59.3 $\pm$ 0.3 & 48.3 & \multirow{2}{*}{7.0\%} \\
        
        + LIP & 6.7 $\pm$ 0.8 $\bullet$ & 70.9 $\pm$ 1.9 $\bullet$ & 51.3 $\pm$ 1.4 $\bullet$ & 64.4 $\pm$ 1.9 $\bullet$ & 53.5 $\pm$ 0.5 $\bullet$ & 63.4 $\pm$ 0.4 $\bullet$ & \textbf{51.7} \\ \midrule

        PL-LEAF & 6.4 $\pm$ 0.8 & 65.1 $\pm$ 2.9 & 45.7 $\pm$ 2.1 & 62.6 $\pm$ 0.9 & 51.1 $\pm$ 0.6 & 60.0 $\pm$ 0.3 & 48.5 & \multirow{2}{*}{3.9\%} \\
        + LIP & 7.3 $\pm$ 0.9 $\bullet$ & 69.4 $\pm$ 2.3 $\bullet$ & 48.3 $\pm$ 2.1 $\bullet$ & 63.6 $\pm$ 0.8 $\bullet$ & 51.8 $\pm$ 0.6 $\bullet$ & 61.7 $\pm$ 1.0 $\bullet$ & \textbf{50.4} \\ 
         \bottomrule
    \end{tabular}
    }       
    \label{tab:real world statistic}
\end{table*}

\begin{table*}[ht!]
\setlength{\tabcolsep}{4.3mm}
\renewcommand\arraystretch{0.76} 
    \centering
    \caption{Classification accuracy(\%) on CIFAR-100N. Experiments are conducted under asymmetric and symmetric noise conditions.}
    \resizebox{\textwidth}{!}{
    \begin{tabular}{c | l l l l | l l l l}\toprule
    
         \multirow{2}{*}{Approaches} & \multicolumn{4}{c|}{Asymmetric} & \multicolumn{4}{c}{Symmetric} 
         
         \\
         & \multicolumn{1}{c}{ $10\%$} & \multicolumn{1}{c}{$20\%$} & \multicolumn{1}{c}{$40\%$} & \multicolumn{1}{c|}{$50\%$} & \multicolumn{1}{c}{$10\%$} & \multicolumn{1}{c}{$20\%$} & \multicolumn{1}{c}{$40\%$} & \multicolumn{1}{c}{$80\%$}
         
         \\ 
         
         \midrule
          PLS & 79.54 & 73.90 & 53.17 & 34.51  & 79.65 & 79.13 & 77.27 & 45.92 
          \\

        + LIP & 79.75 $\bullet$ & 74.34 $\bullet$ & 54.42 $\bullet$ & 38.49 $\bullet$ & 79.87 $\bullet$ & 79.69 $\bullet$ & 77.38 $\bullet$ & 47.43 $\bullet$
        \\ 

        \midrule

        AGCE & 67.32 & 63.57 & 47.25 & 27.90 & 67.61 & 64.75 & 59.78 & 24.05 \\
        + LIP & 67.56 $\bullet$ & 63.99 $\bullet$ & 48.79 $\bullet$ & 28.25 $\bullet$ & 67.98 $\bullet$ & 64.93 $\bullet$ & 59.91 $\bullet$ & 25.43 $\bullet$ \\

         
        

         \bottomrule
    \end{tabular} 
    }
    
    \label{tab:NLL statistic}
\end{table*}

\begin{table*}[ht!]
\renewcommand\arraystretch{1} 
    \centering
    \caption{Classification accuracy on CIFAR-100 and CUB-200. ``Clean'' represents the performance learnt from clean datasets.}
    \resizebox{\textwidth}{!}{
    \begin{tabular}{c l l l l l l }\toprule
    
         \multirow{2}{*}{Approaches} & \multicolumn{3}{c}{CIFAR-100} & \multicolumn{3}{c}{CUB-200}\\\cline{2-7} &\multicolumn{1}{c}{ $p=0.05$} & \multicolumn{1}{c}{$p=0.1$} & \multicolumn{1}{c}{$p=0.2$} & \multicolumn{1}{c}{$p=0.02$} & \multicolumn{1}{c}{$p=0.04$} &
         \multicolumn{1}{c}{$p=0.06$} \\ 
         
         \midrule
         
        PICO & 73.51 $\pm$ 0.22\% & 72.18 $\pm$ 0.32\% & 71.62 $\pm$ 0.18\% & 72.56 $\pm$ 0.07\% & 72.27 $\pm$ 0.46\% & 71.88 $\pm$ 0.33\% \\
        
        + LIP & 73.83 $\pm$ 0.04\% $\bullet$ & 72.43 $\pm$ 0.21\% $\bullet$ & 71.98 $\pm$ 0.11\% $\bullet$ & 72.90 $\pm$ 0.04\% $\bullet$ & 72.56 $\pm$ 0.02\% $\bullet$ & 72.26 $\pm$ 0.03\% $\bullet$ \\ \rowcolor{gray!20} \hline
        
        Clean & \multicolumn{3}{c}{Clean: 73.88 $\pm$ 0.07 \% \quad Clean + LIP: 74.09 $\pm$ 0.03\% $\bullet$} & \multicolumn{3}{c}{Clean: 76.00 $\pm$ 0.34\% \quad Clean + LIP: 76.56 $\pm$ 0.06\% $\bullet$} \\
        
        \midrule
        
        PRODEN & 77.12 $\pm$ 0.13\% & 76.04 $\pm$ 0.16\% & 56.83 $\pm$ 0.05\% & 74.53 $\pm$ 0.03\% & 74.36 $\pm$ 0.15\% & 72.01 $\pm$ 0.18\% \\

        + LIP & 77.25 $\pm$ 0.03\% $\bullet$ & 76.42 $\pm$ 0.05\% $\bullet$ & 56.90 $\pm$ 0.01\% $\bullet$ & 74.73 $\pm$ 0.01\% $\bullet$ & 74.51 $\pm$ 0.02\% $\bullet$ & 72.33 $\pm$ 0.03\% $\bullet$ \\ \rowcolor{gray!20} \hline
        
        Fully Supervised & \multicolumn{3}{c}{Clean: 77.31 $\pm$ 0.07\% \quad Clean + LIP: 77.48 $\pm$ 0.14\% $\bullet$} & \multicolumn{3}{c}{Clean: 75.70 $\pm$ 0.10\% \quad Clean + LIP: 75.91 $\pm$ 0.03\% $\bullet$}\\
        
         \bottomrule
         
    \end{tabular} }  
    \label{tab:deep-learning statistic}
\end{table*}

\begin{table*}[ht]
\setlength{\tabcolsep}{3.7mm}
\renewcommand\arraystretch{0.6} 
    \centering
    \caption{Ablation study of LIP coupled with PL-CL.}
    \resizebox{\textwidth}{!}{
    \begin{tabular}{c c c c c c c c c }\toprule
        \multirow{2}{*}{PSP} & \multirow{2}{*}{LAP} & \multicolumn{6}{c}{Data set} & \multirow{2}{*}{Avg.}\\ \cline{3-8} & & \multicolumn{1}{c}{FG-NET*} & \multicolumn{1}{c}{Lost*} & \multicolumn{1}{c}{MSRCv2*} & \multicolumn{1}{c}{Mirflickr*} & \multicolumn{1}{c}{Soccer Player*} & \multicolumn{1}{c}{Yahoo!News*} \\ \midrule

          \ding{53} & \ding{53} &  6.6 $\pm$ 1.2 & 68.3 $\pm$ 2.4 & 45.1 $\pm$ 1.5 & 63.2 $\pm$ 1.3 & 52.1 $\pm$ 0.5 & 60.9 $\pm$ 0.4 & 49.4 \\ 

          \checkmark & \ding{53} & 6.8 $\pm$ 1.0 & 71.9 $\pm$ 1.9 & 50.2 $\pm$ 1.6 & 63.1 $\pm$ 1.4 & 52.5 $\pm$ 0.5 & 61.7 $\pm$ 0.2 & 51.0 \\

         \checkmark & \checkmark & 6.9 $\pm$ 1.0  & 72.8 $\pm$ 1.6  & 52.1 $\pm$ 2.0  & 64.1 $\pm$ 1.6  & 53.6 $\pm$ 0.5  & 63.9 $\pm$ 0.3 & {52.2}\\
       
         \bottomrule
    \end{tabular}
    }       
    \label{tab:ablation study}
\end{table*}

\subsection{Complexity Analysis}\label{sec complexity}
We here present the computational complexity of LIP. 
Specifically, PSP focuses on performing SVD
on the weights and retaining the principal components, with the complexity of $O(\min(q^2l, ql^2))$. LAP focuses on refining the singular values, with the complexity of $O(nql)$.
Therefore, the overall computational complexity is $O(nql + \min(q^2l, ql^2))$.
In practice, training a classifier like PRODEN requires several hours. 
\textbf{In contrast, LIP can achieve a substantial performance improvement in just one second.}
For example, the execution times for CIFAR-
100 and CUB-200 are 9.98 ms and 16.13 ms, respectively.

%% file: sec/5.experiments.tex
\section{Experiments}

\subsection{Baselines}

To evaluate the effectiveness of LIP, we integrate it with a variety of approaches including PL-CL \cite{jia2023complementary}, PL-CGR \cite{zhang2024partial}, DPCLS \cite{jia2023partial}, PL-AGGD \cite{wang2021adaptivePLAGGD}, SURE \cite{feng2019partialsssfagdsg}, LALO \cite{2018Leveraginglalo}, PL-LEAF \cite{zhang2016partial23232}, PRODEN \cite{lv2020progressiveproden}, PICO \cite{wang2022pico}, PLS \cite{albert2023your} and AGCE \cite{zhou2023asymmetric}.
For more details, please refer to the appendix.

\subsection{Datasets and Settings}
We considered various dataset configurations to explore different forms of label inaccuracy.
First, we evaluated six real-world partial label datasets collected from diverse domains and tasks, including FG-NET \cite{panis2016overviewFg-net}, Lost \cite{cour2009learningLost}, Soccer Player \cite{zeng2013learningSoccerplayer}, Yahoo!News \cite{guillaumin2010multipleYahoonews}, MSRCv2 \cite{liu2012conditional}, and Mirflickr \cite{huiskes2008mirMirflickr}.
In these datasets, each sample, in addition to its correct label, was also assigned several other labels as ``ground-truth'' labels.
Building on this, we further increased the difficulty by randomly selecting 10\% of the samples and completely removing their correct labels, thereby creating an even more challenging dataset.

Additionally, we considered two simplified variations of the above setting.
In the first variation, each sample either retains its correct label or is entirely mislabeled.
For this, following \cite{sheng2024adaptive}, we used CIFAR-100N, which contains both asymmetric and symmetric label noise \cite{sun2021webly}.
In the second variation, each sample retains its correct label but also contains additional erroneous labels as ground-truth.
For this setting, we conducted experiments on two benchmark datasets: CIFAR-100 \cite{krizhevsky2009learning} and CUB-200 \cite{welinder2010caltech}.
Following \cite{lv2020progressiveproden, wang2022pico}, we generated false positive labels by flipping negative labels for each sample with a defined probability $p$ and then combining them with the ground-truth labels to form the final candidate label set.
Specifically, for CIFAR-100, $p$ was set to 0.05, 0.1, and 0.2, while for CUB-200, $p$ was set to 0.02, 0.04, and 0.06.
For more details on these datasets, please refer to the Appendix.

\subsection{Implementation Details}

LIP has only one hyper-parameter, $k$. Empirically, setting $k$ to $\lceil 0.8l \rceil$ yields superior performance. However, given the rapid execution speed of LIP shown in Sec. \ref{sec complexity}, we can also conduct a grid search on $k$ over the validation set to identify the optimal parameter, and then use the optimal $k$ on the test set.
Both PICO and PRODEN utilize ResNet-34 as their backbone architecture. For PRODEN, 500 iterations are conducted on both datasets. 
For PICO, 500 iterations are run on CUB-200 and 800 iterations on CIFAR-100.
For CIFAR-100N, we use ResNet-50 as backbone and an SGD optimizer with a momentum of 0.9 for 300 epochs.
All hyper-parameters for baselines are based on their papers. 
We use MATLAB and PyTorch \cite{paszke2019pytorch} to implement all the methods.
Ten runs of 50\%/50\% random train/test splits are performed on real-world datasets, and the average accuracy with standard deviation is represented.
Five runs are performed with the the average accuracy and standard deviation recorded for CIFAR-100 and CUB-200.

\begin{figure*}[!ht]
    \centering
    \includegraphics[width=0.96\linewidth]{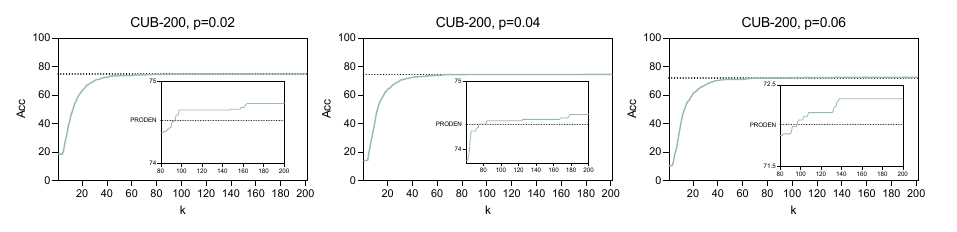}
    \caption{Sensitivity analysis on $k$. LIP is coupled with PRODEN on CUB-200 dataset.}
    \label{fig:sensitivity}
\end{figure*}

\subsection{Results}
Tables \ref{tab:real world statistic}-\ref{tab:deep-learning statistic} present the experimental results of LIP, where we can observe
\begin{itemize}
    \item LIP significantly improves the performance of all these approaches across all cases.
    \item For state-of-the-art methods like PL-CL, PICO, and PLS, LIP serves as a powerful enhancement, further boosting their already impressive performance. For example, on the MSRCv2 dataset, PL-CL experiences a substantial \textbf{15.5\%} improvement, which showcases LIP’s capacity to elevate even the top-tier models.
    \item Other algorithms, such as LALO and PL-LEAF, can achieve even superior performance compared to state-of-the-art methods when coupled with LIP. This suggests that LIP not only enhances established approaches but also unlocks the latent potential of these algorithms. 
    \item For deep learning-based approaches, LIP enables them to approach the performance of models trained on clean, fully supervised data, even in the presence of noisy or incomplete labels. Remarkably, LIP is not only beneficial in weakly supervised scenarios but also in fully supervised settings. LIP can still provide performance gains, further refining the model’s accuracy and generalization.
\end{itemize}

\subsection{Ablation Study}

We conducted an ablation study to assess the contributions of the PSP and LAP modules to the overall performance of LIP. 
Table \ref{tab:ablation study} and Table 2 in the appendix present the experimental results when LIP is coupled with PL-CL and PRODEN, respectively.
It is obvious that both PSP and LAP contribute positively to the performance of the model across all datasets.
PSP effectively preserves the information within the principal subspace, while LAP further refines the noise introduced by label ambiguity. 
The interaction between these two modules enhances the model’s ability when handling inaccurate labels.

\subsection{Sensitive Analysis}
We also performed a sensitivity analysis on the parameter $k$ of LIP to assess its impact on performance. 
Specifically, LIP was coupled with PRODEN on the CUB-200 dataset, with the results displayed in Fig. \ref{fig:sensitivity}.
As the value of $k$ increases, the overall performance of the LIP-augmented model consistently improves, eventually stabilizing after a certain threshold. 
Specifically, once $k$ surpasses 100, the combination of LIP with PRODEN consistently delivers superior results compared to PRODEN alone.
This stability of LIP w.r.t. parameter $k$ is a significant advantage, making it highly reliable and robust for practical applications.

%% file: sec/2.related_work.tex
\section{Related Work}

\subsection{Noisy Label Learning}
In real-world applications, perfectly labeled data is often unattainable due to annotation errors, leading to the challenge of noisy label learning (NLL). 
Deep models tend to overfit noisy labels, severely degrading generalization performance. 
To address this, researchers have developed various strategies, including noise-robust training and label noise modeling.
Noise-Robust Training methods enhance model resilience by designing robust loss functions \cite{ghosh2017robust, ma2020normalized, wang2019symmetric}, such as symmetric cross-entropy \cite{wang2019symmetric}, or by dynamically reweighting samples based on model confidence \cite{ren2018learning, huang2019o2u}. 
Early stopping \cite{bai2021understanding} prevents deep models from memorizing label noise, while semi-supervised approaches \cite{berthelot2019mixmatch, nguyen2019self} leverage pseudo-labeling to refine noisy samples.
Label Noise Modeling explicitly estimates noise transition matrices \cite{goldberger2016training, patrini2017making} to correct corrupted labels, while meta-learning approaches \cite{li2019learning, shu2020meta} utilize clean validation sets to optimize noise handling dynamically. Self-supervised strategies such as DivideMix \cite{li2020dividemix} and Co-teaching \cite{han2018co} separate clean and noisy samples, improving training stability. Despite these advances, why models can learn from inaccurate annotations remains mystery.

\subsection{Partial Label Learning}
Partial Label Learning (PLL) is a weakly supervised learning framework where each instance is associated with multiple candidate labels, with only one being the ground truth. The key challenge in PLL is disambiguation \cite{2018Leveraginglalo,feng2019partialsssfagdsg,nguyen2008classificationplsvm}, which can be broadly categorized into averaging-based and identification-based methods.
Averaging-based methods \cite{cour2011learning,hullermeier2006learningPL-KNN} assume equal contributions from all candidate labels, averaging their outputs for prediction. For instance, PL-KNN \cite{hullermeier2006learningPL-KNN} estimates ground truth by leveraging neighboring instances’ label confidence but suffers from susceptibility to false positives. In contrast, identification-based methods \cite{feng2019partialsssfagdsg,wang2021adaptivePLAGGD} treat the ground truth as a latent variable, refining its estimation via Expectation-Maximization (EM). PL-AGGD \cite{wang2021adaptivePLAGGD} employs feature manifold structures for label confidence estimation, while PL-CL utilizes a complementary classifier to assist disambiguation.
Recent deep learning approaches have further advanced PLL \cite{lv2020progressiveproden,xu2021instancevalen,he2022partialplls,wu2022revisiting,lyu2022deep}. PICO \cite{wang2022pico} resolves label ambiguity via contrastive learning with embedding prototypes, while PRODEN \cite{lv2020progressiveproden} jointly updates the model and refines label identification. Additionally, \cite{he2022partialplls} employs semantic label representations with weighted calibration rank loss to improve disambiguation.

%% file: sec/6.conclusion.tex
\section{Conclusion}

In this paper, we systematically investigate why models can extract correct label information from inaccurate annotations.
By analyzing the weight matrix used for classification, both empirically and theoretically, we reveal that within a certain range, label inaccuracy does not significantly alter the principal subspace of the weight matrix, and in some cases, even leaves it unchanged. 
This observation suggests that the core structure encoding task-relevant information remains robust to moderate levels of label corruption, enabling effective learning despite annotation errors.
Building on these insights, we introduce LIP, a lightweight post-processing plug-in that requires no additional training yet significantly enhances the performance of existing methods. 
By selectively refining model outputs while preserving critical subspace information, LIP provides a simple yet effective solution to mitigating the impact of label inaccuracy. Our extensive evaluations confirm its effectiveness in improving classification performance.
We hope that this work not only deepens the understanding of why models can learn from inaccurate annotations but also inspires new strategies for enhancing robustness in weakly supervised learning paradigms.